\documentclass[sn-basic, iicol, Numbered]{sn-jnl}

\usepackage{array}
\usepackage{arydshln}
\usepackage{graphicx}
\usepackage{amsmath,amssymb,amsfonts}%
\usepackage{amsthm}%
\usepackage{mathrsfs}%
\usepackage[title]{appendix}%
\usepackage{xcolor}%
\usepackage{textcomp}%
\usepackage{manyfoot}%
\usepackage{algorithm}%
\usepackage{algorithmicx}%
\usepackage{algpseudocode}%
\usepackage{listings}%
\usepackage{dblfloatfix}
\usepackage{ulem}
\usepackage{booktabs}
\usepackage{float}
\usepackage{multirow}
\usepackage{tabulary}
\usepackage{placeins}
\setcitestyle{authoryear}
\setcitestyle{round}

\definecolor{codegreen}{rgb}{0,0.6,0}
\definecolor{codegray}{rgb}{0.5,0.5,0.5}
\definecolor{codepurple}{rgb}{0.58,0,0.82}
\definecolor{backcolour}{rgb}{0.95,0.95,0.92}
\usepackage{color, colortbl}

\newcommand{\white}[1]{\textcolor[rgb]{1,1,1}{#1}}

\makeatletter
\@fpsep\textheight
\makeatother

\begin{document}

\title[Article Title]{PanAf20K: A Large Video Dataset for Wild Ape Detection and Behaviour Recognition}

\author*[1]{\fnm{Otto} \sur{Brookes}}\email{otto.brookes@bristol.ac.uk}
\author*[1]{\fnm{Majid} \sur{Mirmehdi}}\email{m.mirmehdi@bristol.ac.uk}
\author[2]{\fnm{Colleen} \sur{Stephens}}
\author[2]{\fnm{Samuel} \sur{Angedakin}}
\author[2]{\fnm{Katherine} \sur{Corogenes}}
\author[3]{\fnm{Dervla} \sur{Dowd}}
\author[2]{\fnm{Paula} \sur{Dieguez}}
\author[6]{\fnm{Thurston C.} \sur{Hicks}}
\author[2]{\fnm{Sorrel} \sur{Jones}}
\author[2]{\fnm{Kevin} \sur{Lee}}
\author[2,3]{\fnm{Vera} \sur{Leinert}}
\author[2]{\fnm{Juan} \sur{Lapuente}}
\author[2]{\fnm{Maureen S.} \sur{McCarthy}}
\author[2]{\fnm{Amelia} \sur{Meier}}
\author[2]{\fnm{Mizuki} \sur{Murai}}
\author[3]{\fnm{Emmanuelle} \sur{Normand}}
\author[3]{\fnm{Virginie} \sur{Vergnes}}
\author[4]{\fnm{Erin G.} \sur{Wessling}}
\author[2,7]{\fnm{Roman M.} \sur{Wittig}}
\author[8]{\fnm{Kevin} \sur{Langergraber}}
\author[2]{\fnm{Nuria} \sur{Maldonado}}
\author[1]{\fnm{Xinyu} \sur{Yang}}
\author[5]{\fnm{Klaus} \sur{Zuberbühler}}
\author[2,3]{\fnm{Christophe} \sur{Boesch}}
\author*[2]{\fnm{Mimi} \sur{Arandjelovic}}
\author*[2,9]{\fnm{Hjalmar} \sur{Kühl}}\email{hjalmar.kuehl@senckenberg.de}
\author*[1]{\fnm{Tilo} \sur{Burghardt}}\email{tilo@cs.bris.ac.uk
}

\affil*[1]{\orgdiv{Department of Computer Science}, \orgname{University of Bristol}, \orgaddress{\country{United Kingdom}}}

\affil[2]{\orgdiv{Max Planck Institute for Evolutionary Anthropology}, \orgaddress{\city{Leipzig}, \country{Germany}}}

\affil[3]{\orgname{Wild Chimpanzee Foundation}, \orgaddress{\city{Leipzig}, \country{Germany}}}

\affil[4]{\orgdiv{Department of Human Evolutionary Biology}, \orgname{Harvard University}, \orgaddress{\state{Massachusetts}, \country{USA}}}

\affil[5]{\orgdiv{School of Psychology \& Neuroscience}, \orgname{University of St Andrews}, \orgaddress{\city{St. Andrews}, \country{Scotland}}}

\affil[6]{\orgdiv{Faculty of Artes Liberales}, \orgname{University of Warsaw}, \orgaddress{\city{Warsaw}, \country{Poland}}}

\affil[7]{\orgdiv{Institute for Cognitive Sciences Marc Jeannerod}, \orgname{University of Lyon}, \orgaddress{\city{Lyon}, \country{France}}}

\affil[8]{\orgdiv{Institute of Human Origins}, \orgname{Arizona State University}, \orgaddress{\state{Arizona}, \country{USA}}}

\affil[9]{\orgdiv{Senckenberg Museum of Natural History Goerlitz}, \orgaddress{\city{Goerlitz}, \country{Germany}}}


\abstract{We present the PanAf20K dataset, the largest and most diverse open-access annotated video dataset of great apes in their natural environment. It comprises more than 7 million frames across $\sim$20,000 camera trap videos of chimpanzees and gorillas collected at 14 field sites in tropical Africa as part of the Pan African Programme: The Cultured Chimpanzee. The footage is accompanied by a rich set of annotations and benchmarks making it suitable for training and testing a variety of challenging and ecologically important computer vision tasks including ape detection and behaviour recognition. Furthering AI analysis of camera trap information is critical given the International Union for Conservation of Nature now lists all species in the great ape family as either Endangered or Critically Endangered. We hope the dataset can form a solid basis for engagement of the AI community to improve performance, efficiency, and result interpretation in order to support assessments of great ape presence, abundance, distribution, and behaviour and thereby aid conservation efforts. The dataset and code are available from the project website: \href{https://obrookes.github.io/panaf.github.io/}{PanAf20K}}


\maketitle


\section{Introduction}\label{sec:intro}

\begin{figure*}[hbt!]
\centering
\includegraphics[width=0.95\linewidth]{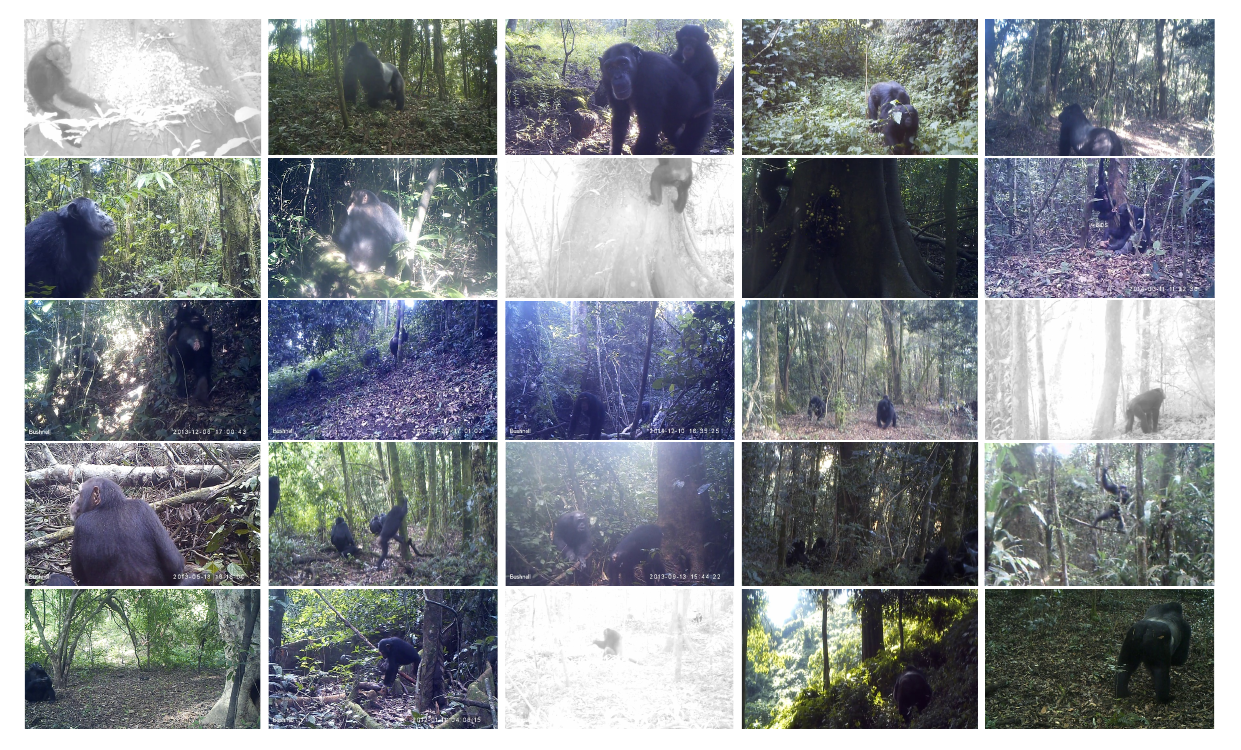}\vspace{-2pt}
\caption{\textbf{PanAf20K Visual Overview}. We present the largest and most diverse open-access video dataset of great apes in the wild. It comprises $\sim$20,000 videos and more than 7 million frames extracted from camera traps at 14 study sites spanning 6 African countries. Shown are 25 representative still frames from the dataset highlighting its diversity with respect to many important aspects such as behavioural activities, species, number of apes, habitat, day/night recordings, scene lighting, and more.}
\label{fig:dataset_overview}
\end{figure*}

\textbf{Motivation.} As the biodiversity crisis intensifies, the survival of many endangered species grows increasingly precarious, evidenced by species diversity continuing to fall at an unprecedented rate~\citep{vie2009wildlife,ceballos2020vertebrates}. The great ape family, whose survival is threatened by habitat degradation and fragmentation, climate change, hunting and disease, is a prime example~\citep{carvalho2021predicting}. The International Union for Conservation of Nature (IUCN) considers all three member species, that is orangutans, gorillas, chimpanzees (including bonobos), to be either endangered or critically endangered. 

The threat to great apes has far-reaching ecological implications. Great apes contribute to the balance of healthy ecosystems by seed dispersal, consumption of leaves and bark, and shaping habitats by creating canopy gaps and trails~\citep{haurez2015role, tarszisz2018ecophysiologically, chappell2022role}. They also form part of complex forest food webs, their removal from which would have cascading consequences for local food chains. In addition, great apes are our closest evolutionary relatives and a key target for anthropological research. We share 97\% of our DNA with the phylogenetically most distant orangutans and 98.8\% with the closer chimpanzees and bonobos. The study of great apes, including their physiology, genetics, and behaviour, is essential to addressing questions of human nature and evolution~\citep{pollen2023human}. Urgent conservation action for the protection and preservation of these emblematic species is therefore essential. 

The timely and efficient assessment of great ape presence, abundance, distribution, and behaviour is becoming increasingly important in evaluating the effectiveness of conservation policies and intervention measures. The potential of exploiting camera trap imagery for conservation or biological modelling is well recognised~\citep{kuehl2013,tuia2022}. However, even small camera networks generate large volumes of data~\citep{fegraus2011data} and the number and complexity of downstream processing tasks required to perform ecological analysis is extensive. Typically, ecologists first need to identify those videos that contain footage of the target study species followed by further downstream analyses, such as estimating the distance of the animals from the camera (i.e., camera trap distance sampling) to calculate species density or identification of ecologically or anthropologically important behaviours, such as tool use or camera reactivity~\citep{houa2022animal}. Performing these tasks manually is time consuming and limited by the availability of human resources and expertise, becoming infeasible at large scale. This underlines the need for rapid, scalable, and efficient deep learning methods for automating the detection and assessment of great ape populations and analysis of their behaviours.

To facilitate the development of methods for automating the interpretation of camera trap data, large-scale, open-access video datasets must be available to the relevant scientific communities, whilst removing geographic details that could potentially threaten the safety of animals~\citep{tuia2022}. Unlike the field of human action recognition and behaviour understanding, where several large, widely acknowledged datasets exist for benchmarking~\citep{kuehne2011hmdb, soomro2012ucf101, kay2017kinetics}, the number of great ape datasets is limited and those that are currently available lack scale, diversity and rich annotations.

\textbf{Contribution.} In this study, we present the PanAf20K dataset, the largest and most diverse open-access video dataset of great apes in the wild -- ready for AI training. The dataset comprises footage collected from 14 study sites across 6 African countries, featuring apes in over 20 distinct habitats (i.e., forests, savannahs, and marshes). It displays great apes in over 100 individual locations (e.g., trails, termite mounds, and water sources) displaying an extensive range of 18 behaviour categories. A visual overview of the dataset is presented in  Fig.~\ref{fig:dataset_overview}. The footage is accompanied by a rich set of annotations suitable for a range of ecologically important tasks such as detection, action localisation, fine-grained and multi-label behaviour recognition.

\textbf{Paper Organisation}. Following this introduction, Sec.~\ref{sec:related_work} reviews existing animal behaviour datasets and methodologies for great ape detection and behaviour recognition. Sec.~\ref{sec:dataset_overview} describes both parts of the dataset, the PanAf20K and the PanAf500, and details how the data was collected and annotated. Benchmark results for several computer vision tasks are presented in Sec~\ref{sec:benchmarks}. Sec~\ref{sec_discussion} discusses the main findings as well as any limitations alongside future research directions while Sec~\ref{sec:conclusion} summarises the dataset and highlights its potential applications.


\section{Related Work}\label{sec:related_work}

\textbf{Great Ape Video Datasets for AI Development.} While there have been encouraging trends in the creation of new animal datasets~\citep{swanson2015snapshot,cui2018large,van2018inaturalist,beery2021iwildcam}, there is still only a limited number specifically designed for great apes and even fewer suitable for behavioural analysis. In this section, the most relevant datasets are described. 

Bain et al.~\citep{bain2021automated}, curated a large camera trap video dataset ($>40$ hours) with fine-grained annotations for two behaviours; buttress drumming and nut cracking. However, the data and corresponding annotations are not yet publicly available and the range of annotations is limited to two audio-visually distinct behaviours. The Animal Kingdom dataset~\citep{ng2022animal}, created for advancing behavioural understanding, comprises footage sourced from YouTube (50hr, 30K videos) along with annotations that cover a wide range of actions, from eating to fighting. The MammalNet dataset~\citep{chen2023mammalnet}, which is larger and more diverse, is also composed from YouTube footage (18K videos, 539 hours) and focuses on behavioural understanding across species. It comprises taxonomy-guided annotations for 12 common behaviours, identified through previous animal behaviour studies, for 173 mammal categories. While both datasets are valuable resources for the study of animal behaviour, they contain relatively few great ape videos since these species make up only a small proportion of the overall dataset. Animal Kingdom spans $\sim$100 videos while MammalNet includes $\sim$1000 videos across the whole great ape family, representing $\sim$0.5\% and $\sim$5\% of all videos, respectively. Other work to curate great ape datasets has focused annotation efforts on age, sex, facial location, and individual identification~\citep{freytag2016chimpanzee,brookes2020dataset,schofield2019chimpanzee}, rather than behaviour.

For the study of great ape behaviour, the currently available datasets have many limitations. First, they are too small to capture the full breadth of behavioural diversity. This is particularly relevant for great apes, which are a deeply complex species, displaying a range of individual, paired and group behaviours, that are still not well understood~\citep{tennie2016nature, samuni2021group}. Secondly, they are not composed of footage captured by sensors commonly used in ecological studies, such as camera traps and drones. This means that apes are not observed in their natural environment and the distribution of behaviours will not be representative of the wild (i.e., biased towards 'interesting' or 'entertaining' behaviours). Additionally, the footage may be biased towards captive or human-habituated animals which display altered or unnatural behaviours and are unsuitable for studying their wild counterparts~\citep{clark2011great,chappell2022role}. All these factors may limit the ability of trained models to generalise effectively to wild footage of great apes where conservation efforts are most urgently needed. This, in turn, limits their practical and \textit{immediate} utility. We aim to overcome these limitations by introducing a large scale, open-access video dataset that enables researchers to develop models for analysing the behaviour of great apes in the wild and evaluate them against established methods.\white{xxx}

\textbf{Great Ape Detection \& Individual Recognition}. Yang \textit{et al}.~\citep{yang2019great} developed a multi-frame system capable of accurately detecting the full body location of apes in challenging camera-trap footage. In more recent work, Yang \textit{et al}. developed a curriculum learning approach that enables the utilisation of large volumes of unlabelled data to improve detection performance~\citep{yang2023dynamic}.
Several other works focus on facial detection and individual identification. In early research, Freytag \textit{et al.}~\citep{freytag2016chimpanzee} applied YOLOv2~\citep{redmon2017yolo9000}, to localise the faces of chimpanzees. They utilised a second deep CNN for feature extraction (AlexNet~\citep{krizhevsky2012imagenet} and VGGFaces~\citep{parkhi2015deep}), and a linear support vector machine for identification. Later, Brust \textit{et al.}~\citep{brust2017towards} extended their work utilising a much larger and diverse dataset. Schofield \textit{et al.}~\citep{schofield2019chimpanzee} presented a pipeline for identification of 23 chimpanzees across a video archive spanning 14 years. Similar to~\citep{brust2017towards}, they trained the single-shot object detector, SSD~\citep{schofield2019chimpanzee}, to perform initial localisation, and a secondary CNN model to perform individual classification. Brookes \textit{et al}.~\citep{brookes2020dataset} employed YOLOv3~\citep{redmon2018yolov3} to perform \textit{one-step} simultaneous facial detection and individual identification on captive gorillas.

\textbf{Great Ape Action \& Behaviour Recognition}. To date, three systems have attempted automated great ape behavioural action recognition. The first~\citep{sakib2020visual} was based on the two-stream convolutional architecture by ~\citep{simonyan2014two} and uses 3D ResNet-18s for feature extraction and LSTM-based fusion of RGB and optical flow features. They reported a strong top-1 accuracy of 73\% across the nine behavioural actions alongside a relatively low average per class accuracy of 42\%. The second, proposed by Bain \textit{et al}.~\citep{bain2021automated}, utilises both audio and video inputs to detect two specific behaviours; buttress drumming and nut cracking. Their system utilises a 3D ResNet-18 and a 2D ResNet-18 for extraction of visual and audio features, respectively, in different streams. They achieved an average precision of 87\% for buttress drumming and 85\% for nut cracking on their unpublished dataset. Lastly, Brookes \textit{et al}.~\citep{brookes2023triple} introduced a triple-stream model that utilises RGB, optical flow and DensePose within a metric learning framework, and achieved top-1 and average per-class accuracy of 85\% and 65\%, respectively.


\section{Dataset Overview}\label{sec:dataset_overview}

\begin{figure*}[!hb]
\centering
\includegraphics[width=0.95\linewidth]{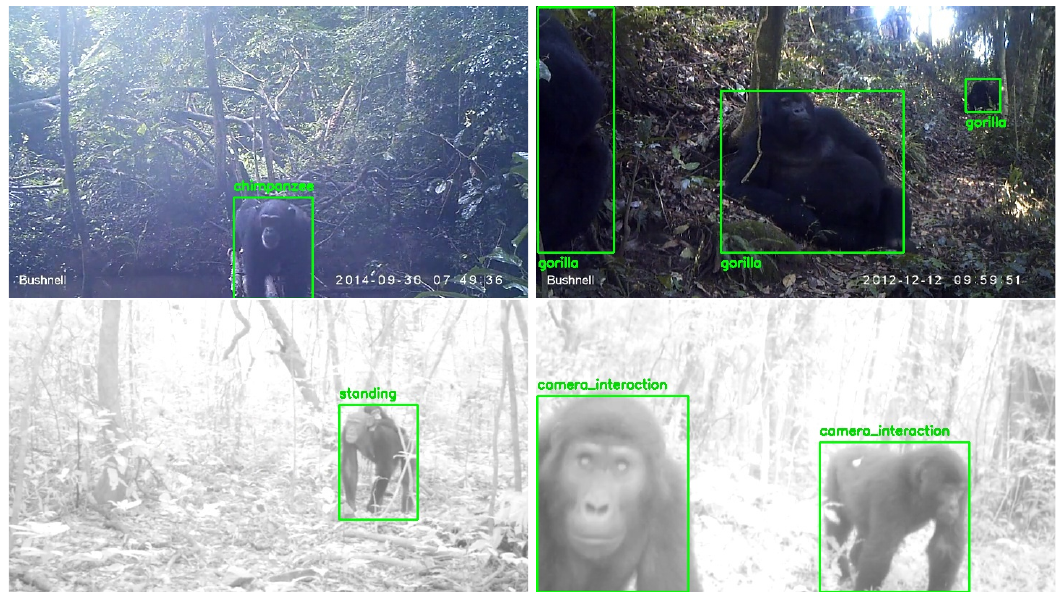}\vspace{-0pt}
\caption{\textbf{Manually annotated full-body location, species and behavioural action labels}. Sample frames extracted from PanAf20K videos with species (row 1) and behavioural action annotations (row 2) displayed. Green bounding boxes indicate the full-body location of an ape. Species and behavioural action annotations are shown in the corresponding text.}
\label{fig:species_behaviour_annotations}
\end{figure*}

\textbf{Task-focused Data Preparation.} The PanAf20K dataset consists of two distinct parts. The first includes a large video dataset containing 19,973 videos annotated with multi-label behavioural labels. The second part comprises 500 videos with fine-grained annotations across $\sim$180,000 frames. Videos are recorded at 24 FPS and resolutions of $720\times404$ for 15 seconds ($\sim$360 frames). In this section, we provide an overview of the dataset, including how the video data was originally collected (see Sec.~\ref{sec:data_collection}) and annotated for both parts (see Sec.~\ref{sec:data_ann}).


\subsection{Data Acquisition}
\label{sec:data_collection}
\textbf{Camera Trapping in the Wild.} The PanAf Programme: The Cultured Chimpanzee has 39 research sites and data collection has been ongoing since January 2010. The data included in this paper samples 14 of these sites and the available data were obtained from studies of varying duration (7–22 months). Grids comprising 20 to 96 $1\times1$ km cells were established for the distribution of sampling units (to cover a minimum of 20–50 km$^2$ in rainforest and 50–100 km$^2$ in woodland savannah). An average of 29 (range 5–41) movement-triggered Bushnell cameras were installed per site. One camera was installed per grid cell where possible. However, in larger grids cameras were placed in alternate cells. If certain grid cells did not contain suitable habitat, such as grassland in forest-savanna mosaic sites, two cameras were placed instead as far away from each other as possible, in cells containing suitable habitat to maximize coverage. In areas where activities of interest (e.g., termite fishing sites) were likely to take place, a second camera was installed to capture the same scene from a different angle. Cameras were placed approx. 1m high above ground, in locations that were frequently used by apes (e.g., trail, fruit trees). This method ensured a strategic installation of cameras, with maximal chance of capturing footage of terrestrial activity of apes. Both GPS location and habitat type for each location was noted. Footage was recorded for 60 seconds with a 1 second interval between triggers and cameras were visited every 1-3 months for maintenance and to download the recorded footage throughout the study periods.


\subsection{Data Annotation}\label{sec:data_ann}

\begin{figure*}[!hb]
\centering
\includegraphics[width=0.95\linewidth]{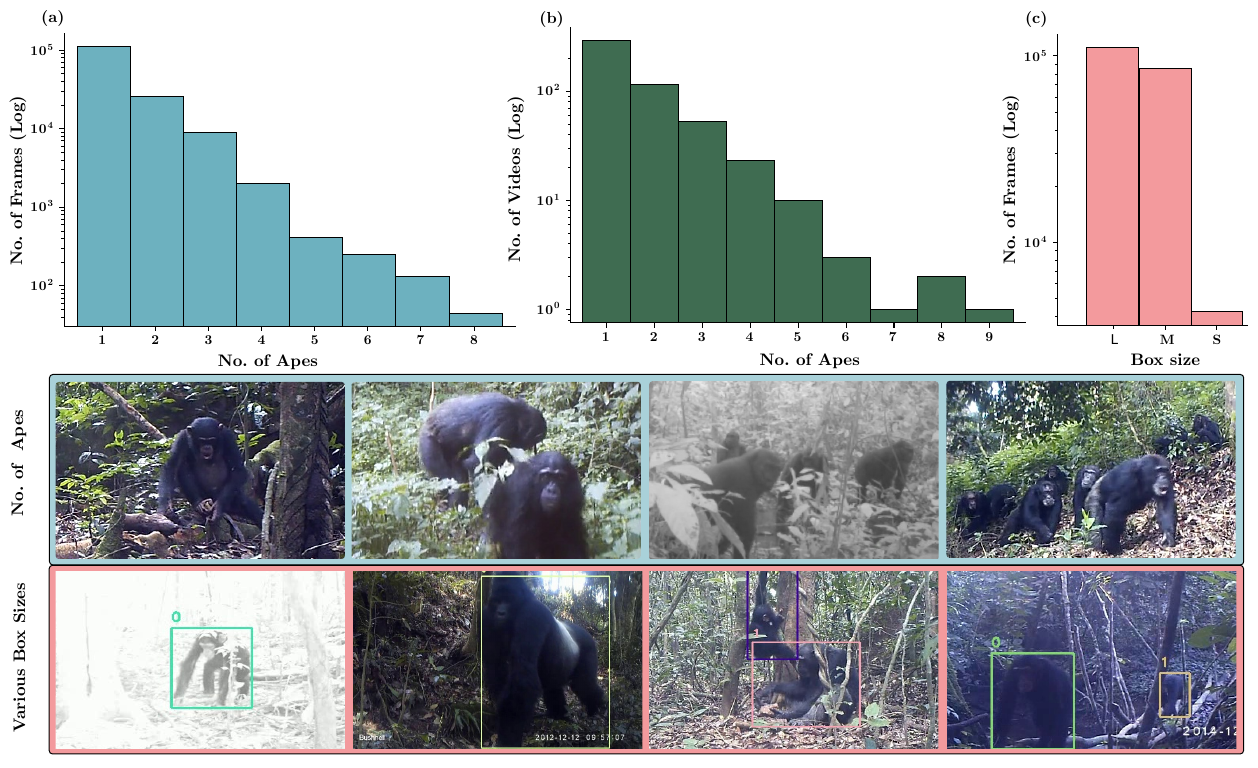}\vspace{-2pt}
\caption{\textbf{Number of Apes \& Bounding Box Size Distribution in the PanAf500 Data}. \textmd{The top row shows the distribution of apes across frames and videos in (a) and (b), respectively, while the distribution of bounding box sizes is shown in (c). The middle row shows still frame examples of videos containing one, two, four and eight apes (viewing from left to right). The bottom row demonstrates still frames with bounding boxes of various sizes; the colour of bounding box and associated number represent the intra-video individual IDs.}}
\label{fig:apes_per_frame}
\end{figure*}

\textbf{Fine-grained Annotation of PanAf500}. 
The PanAf500 was ground-truth labelled by users on the community science platform Chimp\&See~\citep{arandjelovic2016chimp} and researchers at the University of Bristol~\citep{yang2019great,sakib2020visual} (examples are shown in Fig.~\ref{fig:species_behaviour_annotations}). We re-formatted the metadata from these sources specifically for use in computer vision under reproducible and comparable benchmarks ready for AI-use. The dataset includes frame-by-frame annotations for full-body location, intra-video individual identification, and nine behavioural actions~\citep{sakib2020visual} across 500 videos and $\sim$180,000 frames.

\begin{figure*}[!ht]
\centering
\includegraphics[width=0.95\linewidth]{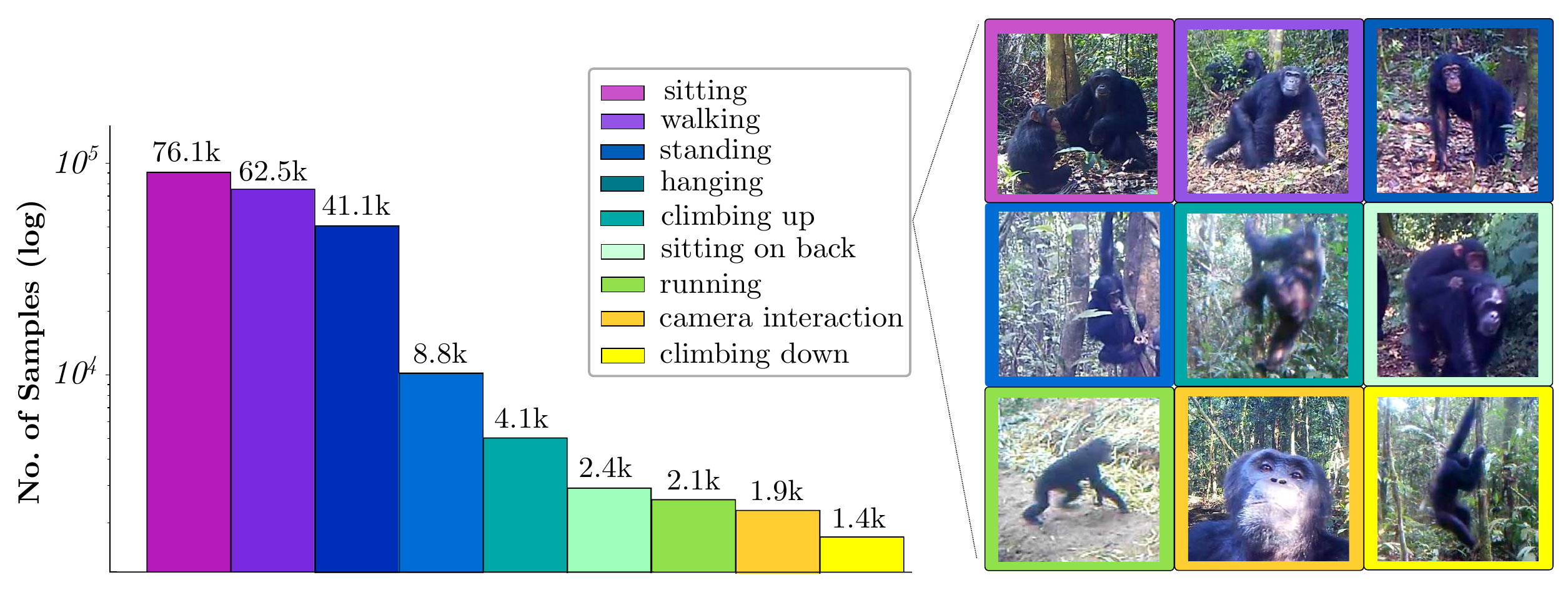}\vspace{-2pt}
\caption{\textbf{Behavioural Actions in the PanAf500 Data}. \textmd{Examples of each one of the nine behavioural action classes (\textit{right}) and their distribution (\textit{left}) across 500 videos. The total number of per-frame annotations for each behavioural action class is shown on top of the corresponding bar.}}
\label{fig:human_class_dist}
\end{figure*}

As shown in Fig.~\ref{fig:apes_per_frame}, the number of individual apes varies significantly, from one to nine, with up to eight individuals appearing together simultaneously. Individuals and pairs occur the most frequently while groups occur less frequently, particularly those exceeding four apes. Bounding boxes are categorised according to the COCO dataset~\citep{lin2014microsoft} (i.e., $>96^2$, $96^2$ and $32^2$ for large, medium and small, respectively) with small bounding boxes occurring relatively infrequently compared to large and medium boxes.

The behavioural action annotations cover 9 basic behavioural actions; sitting, standing, walking, running, climbing up, climbing down, hanging, sitting on back, and camera interaction. We refer to these classes as behavioural actions in recognition of historical traditions in biological and computer vision disciplines, which would consider them behaviours and actions, respectively. Fig. \ref{fig:human_class_dist} displays the behavioural actions classes in focus together with their per-frame distribution. The class distribution is severely imbalanced, with the majority of samples ($>85\%$) belonging to three \textit{head} classes (i.e., sitting, walking and standing). The remaining behavioural actions are referred to as \textit{tail} classes. The same imbalance is observed at the clip level, as shown in Tab.~\ref{tab:samples_per_video}, although the distribution of classes across clips does not match the per-frame distribution exactly. While behavioural actions with longer durations (i.e., sitting) have more labelled frames, this does not necessarily translate to more clips. For example, there are more clips of walking and standing than sitting, and more clips of climbing up than hanging, although the latter have fewer labelled frames.

\begin{table}[!ht]
\centering
\caption{\textbf{Behavioural Action Class Statistics}. \textmd{The total number of clips for each behavioural action alongside the average duration in seconds and frames.}}
\begin{tabular}{crrr}
\toprule
\textbf{Action} & \textbf{Clips} & \textbf{Time (s)} & \textbf{Frames} \\
\midrule
walking & 747 & $3.49 \pm 2.94$ & 83.69 \\
standing & 366 & $4.77 \pm 4.59$ & 114.57 \\
sitting & 308 & $10.30 \pm 5.51$ & 247.13 \\
climbing up & 81 & $2.11 \pm 1.67$ & 50.59 \\
hanging & 50 & $7.35 \pm 5.24$ & 176.28 \\
climbing down & 35 & $1.73 \pm 1.24$ & 41.57 \\
running & 34 & $2.61 \pm 2.06$ & 62.59 \\
camera interaction & 32 & $2.52 \pm 3.77$ & 60.59 \\
sitting on back & 26 & $3.89 \pm 4.04$ & 93.46 \\
\bottomrule
\end{tabular}
\label{tab:samples_per_video}
\end{table}

\begin{figure*}[p!]
\centering
\includegraphics[width=\linewidth]{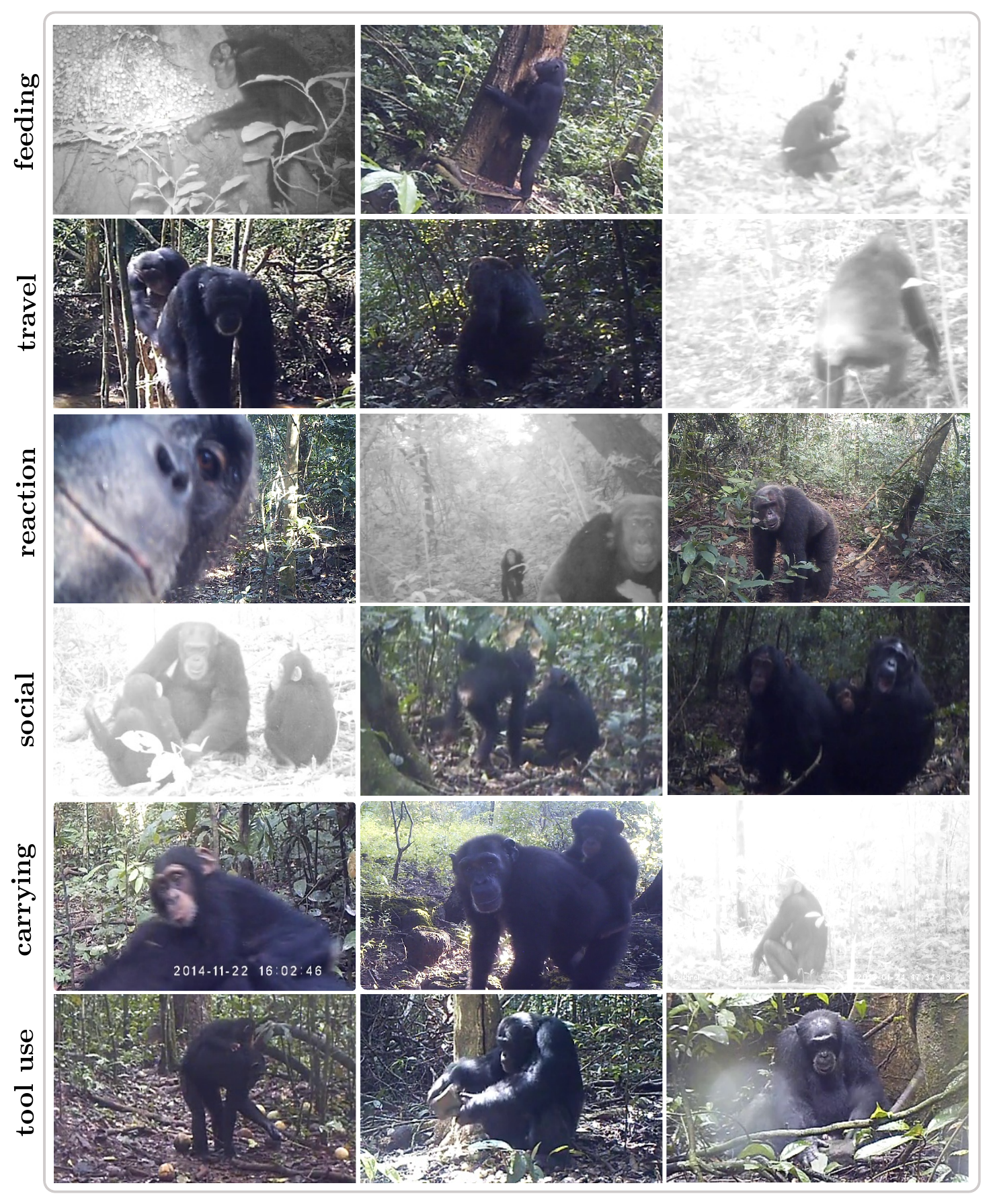}\vspace{-10pt}
\caption{\textmd{\textbf{PanAf20K Behaviour Examples.} Triplets of example frames for six categories (i.e., feeding, travel, camera reaction, social interaction, chimp carrying and tool use) in the PanAf20K dataset are shown. Note that camera reaction, social interaction and chimp carrying have been abbreviated to reaction, social and carrying, respectively.}}
\label{fig:multi_label_examples}
\end{figure*}

\begin{figure*}[!p]
\centering
\includegraphics[width=0.9\linewidth]{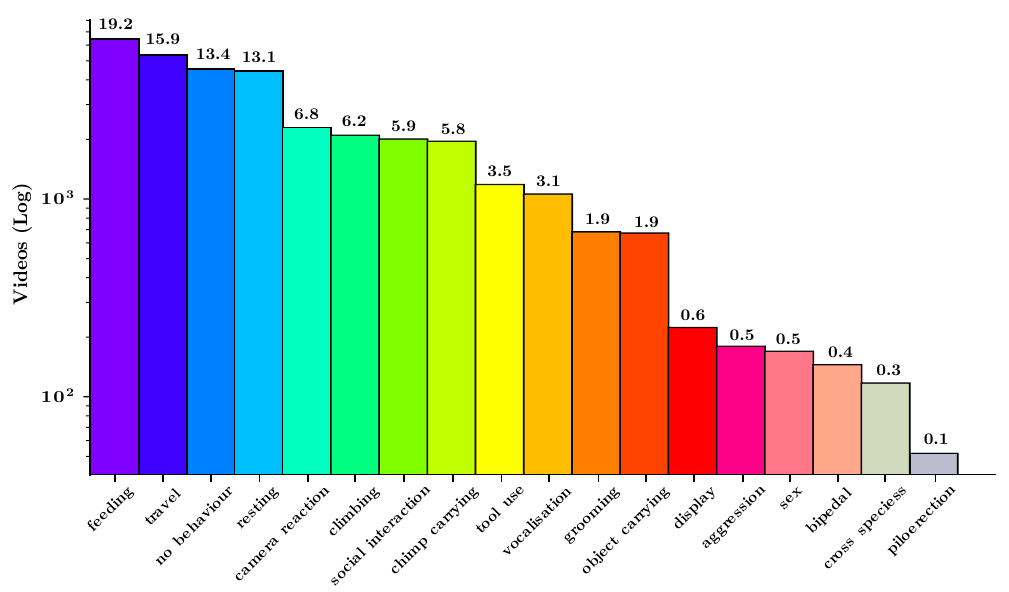}
\caption{\textmd{\textbf{Behavioural Annotations of the PanAf20K Dataset.} The distribution of behaviour categories for the PanAf20K dataset is shown. Figures above each bar represent the dataset proportion (\%) of each class.}}
\label{fig:multi_label_bar}
\bigskip
\includegraphics[width=0.9\linewidth]{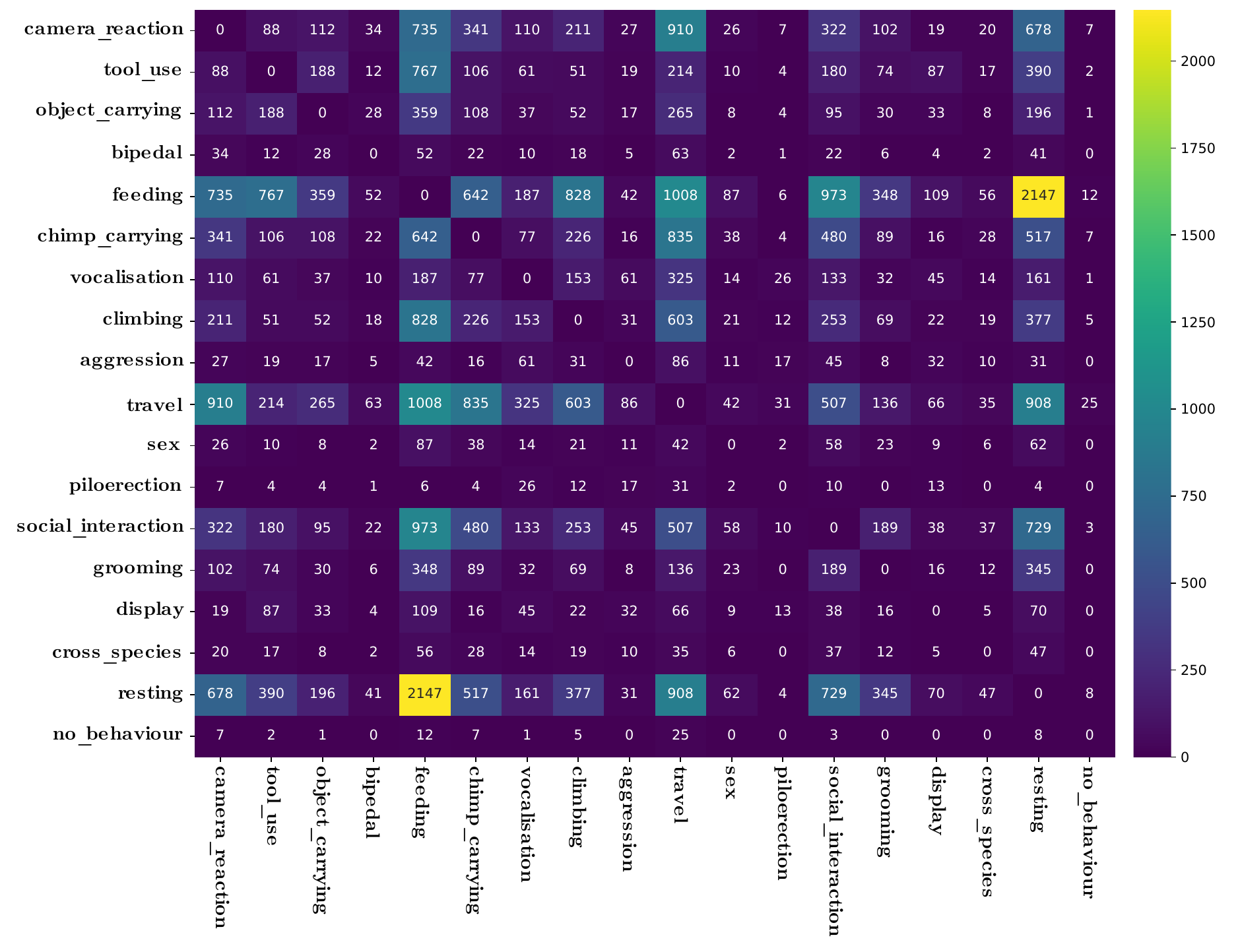}
\caption{\textmd{\textbf{Co-occurrence of Behaviours in the PanAf20k Dataset.} A co-occurrence matrix for the PanAf20K behaviours, where each cell reflects the number of times two behaviours occurred together. Diagonal cells are reset to aid visibility.}}
\label{fig:multi_label_cooc}
\end{figure*}

\begin{figure*}[!ht]
\centering
\includegraphics[width=\linewidth]{Fig8.pdf}\vspace{-10pt}
\caption{\textmd{\textbf{Examples of Fine-grained and Multi-label Annotations}. For videos with fine-grained annotations, full-body locations and behavioural actions are associated with each ape on a frame-by-frame basis (left). In contrast, multi-label behaviour annotations are provided at the video level (right); behaviours are not localised or assigned specifically to each ape.}}
\label{fig:fine_vs_multi}
\end{figure*}

\textbf{Multi-label Behavioural Annotation of PanAf20K}. Community scientists on the Chimp\&See platform provided multi-label behavioural annotations for $\sim$20,000 videos. They were shown 15-second clips at a time and asked to annotate whether animals were present or whether the clip was blank. To obtain a balance between specificity and keeping the task accessible and interesting to a broad group of people, annotators were presented with a choice of classification categories. These categories allowed focus to be given to ecologically important behaviours such as tool use, camera reaction and bipedalism. Hashtags for behaviours not listed in the classification categories were also permitted, allowing new and interesting behaviours to be added when they were discovered in the videos. The new behaviours were subcategories of the existing behaviours, many of them relating to tool use (e.g., algae scooping and termite fishing in aboreal nests).

To ensure annotation quality and consistency a video was only deemed to be analyzed when either three volunteers marked the video as blank, unanimous agreement between seven volunteers was observed, or 15 volunteers annotated the video. These annotations were then extracted and expertly grouped into 18 co-occurring classes, which form the multi-label behavioural annotations presented here. The annotations follow a multi-hot binary format that indicates the presence of one or many behaviours. It should also be noted that behaviours are not assigned to individual apes or temporally localised within each video. Fig.~\ref{fig:multi_label_examples} presents examples for several of the most commonly occurring behaviours. Fig.~\ref{fig:multi_label_bar} shows the full distribution of behaviours across videos, which is highly imbalanced. Four of the most commonly occurring classes are observed in $> 60\%$ videos, while the least commonly occurring classes are observed in $<1\%$. The relationship between behaviours is shown in Fig.~\ref{fig:multi_label_cooc} which presents co-occurring classes. The behaviours differ from the behavioural actions included in the PanAf500 dataset, corresponding to higher order behaviours that are commonly monitored in ecological studies. For example, instances of travel refer to videos that contain an individual or group of apes travelling, whereas associated behavioural actions such as walking or running may occur in many contexts (i.e., walking towards another ape during a social interaction or while searching for a tool).

Both parts of the dataset are suitable for different computer vision tasks. The PanAf500 supports great ape detection, tracking, action grounding, and multi-class action recognition, while the PanAf20k supports multi-label behaviour recognition. The difference between the two annotation types can be observed in Fig.~\ref{fig:fine_vs_multi}.

\textbf{Machine Labels for Animal Location and IDs}. We generated full-body bounding boxes for apes present in the remaining, unlabelled videos using state-of-the-art (SOTA) object detection models evaluated on the PanAf500 dataset (see Sec.~\ref{sec:benchmarks}). Additionally, we assigned intra-video IDs to detected apes using the multi-object tracker, OC-SORT~\citep{cao2022observation}. Note that these pseudo-labels do not yet associate behaviours with individual bounding boxes.


\section{Experiments \& Results}\label{sec:benchmarks} 

\begin{table*}[!ht]
\centering
\caption{\textbf{Ape Detection Benchmarks.} \textmd{Detection performance on the PanAf500 dataset. Results are reported for the MegaDetector~\citep{beery2019efficient}, ResNet-101~(+SCM+TCM)~\citep{yang2019great}, VarifocalNet~\citep{zhang2021varifocalnet}, Swin Transformer~\citep{liu2021swin} and ConvNeXt~\citep{liu2022convnet}. The highest scores for each metric are shown in bold.}}
\centering
\begin{tabular}{crrrrrrr} 
\toprule
\multirow{2}{*}{\textbf{Model}}
& \multicolumn{4}{c}{\textbf{mAP (\%)}} & \multicolumn{3}{c}{\textbf{Other (\%)}} \\
\cmidrule(r){2-5}\cmidrule(l){6-8}
& \multicolumn{1}{c}{All} & \multicolumn{1}{c}{L} & \multicolumn{1}{c}{M} & \multicolumn{1}{c}{S} & \multicolumn{1}{c}{Precision} & \multicolumn{1}{c}{Recall} & \multicolumn{1}{c}{F1} \\
\midrule
\multicolumn{1}{l}{MegaDetector} & \textbf{88.0} & \textbf{98.05} & 82.60 & 68.21 & 56.93 & 90.58 & 69.92  \\
\multicolumn{1}{l}{ResNet-101} & 81.2 & 77.60 & 88.97 & \textbf{88.84} & 42.37 & 88.93 & 57.40 \\
\multicolumn{1}{l}{VarifocalNet} & 84.1 & 84.50 & 88.73 & 82.07 & 21.73 & 88.57 & 34.90 \\
\multicolumn{1}{l}{Swin Transformer} & 87.2 & 82.47 & \textbf{96.86} & 88.53 & \textbf{83.66} & \textbf{92.03} & \textbf{87.65} \\
\multicolumn{1}{l}{ConvNeXt} & 86.6 & 83.51 & 95.16 & 81.13 & 81.80 & 91.93 & 86.57 \\
\bottomrule
\end{tabular}
\label{tab:ape_detection_results_test}
\end{table*}

This section describes experiments relating to the PanAf500 and PanAf20K datasets. For the former, we present benchmark results for great ape detection and fine-grained action recognition. For the latter, we present benchmark results for multi-label behavioural classification. For both sets of experiments, several SOTA architectures are used.


\subsection{PanAf500 Dataset}
\textbf{Baseline Models.} We report benchmark results for ape detection and fine-grained behavioural action recognition for the PanAf500 dataset, trained and evaluated on SOTA architectures. For ape detection, this entails the MegaDetector~\citep{beery2019efficient}, ResNet-101~(+SCM+TCM)~\citep{yang2019great}, VarifocalNet (VFNet)~\citep{zhang2021varifocalnet}, SwinTransformer~\citep{liu2021swin} and ConvNext~\citep{liu2022convnet} architectures. For fine-grained action recognition, we considered X3D~\citep{feichtenhofer2020x3d}, I3D~\citep{carreira2017quo}, 3D ResNet-50~\citep{du2017closer}, Timesformer~\citep{bertasius2021space} and MViTv2~\citep{li2022mvitv2} architectures. Action recognition models were chosen based on SOTA performance on human action recognition datasets and to be consistent with the best performing models on the AnimalKingdom~\citep{ng2022animal} and MammalNet datasets~\citep{chen2023mammalnet}. In all cases, train-val-test (80:05:15) splits were generated at the video-level to ensure generalisation across video/habitat and splits remained consistent across tasks.

\begin{figure*}[!hb]
\begin{minipage}{0.5\textwidth}
    \includegraphics[width=\linewidth]{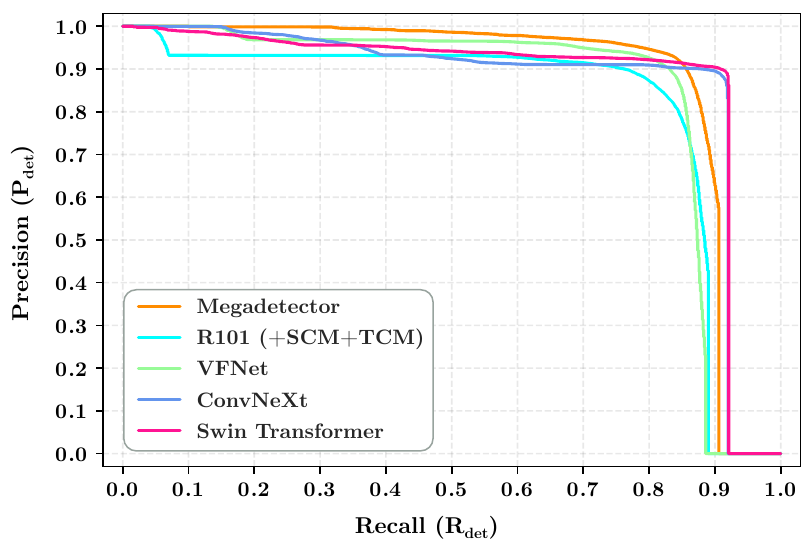}\vspace{-10pt}
    \caption{Megadetector~\citep{beery2019efficient} achieves higher precision for the majority of cases although ConvNeXt~\citep{liu2022convnet} and Swin Transformer~\citep{liu2021swin} achieve better precision scores at high recall rates ($R_{det} > 0.84$). \white{blahblahblahblahblahblahblah}}
    \label{fig:detection_pr_curve}
    \end{minipage}
    \hspace{\fill} 
    \begin{minipage}{0.5\textwidth}
    \includegraphics[width=\linewidth]{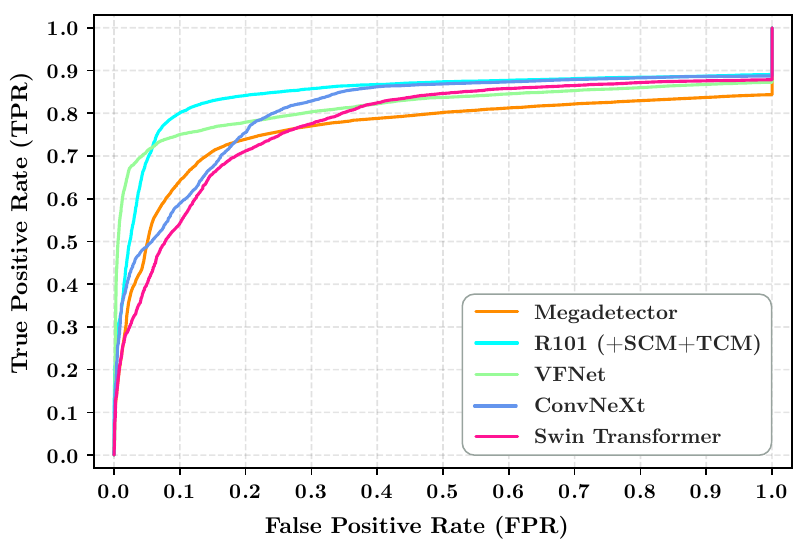}\vspace{-10pt}
    \caption{R101 (+SCM +TCM)~\citep{yang2019great} and VFNet~\citep{zhang2021varifocalnet} achieve the highest true positive rates at low false positive rates ($FPR < 0.15$). At higher false positive rates R101(+SCM+TCM)~\citep{yang2019great} performs better.}
    \label{fig:detection_roc_curve}
    \end{minipage}
\end{figure*}

\textbf{Great Ape Detection}. We initialised all models with pretrained feature extractors. For all models, except the Megadetector, we utilised MS COCO~\citep{lin2014microsoft} pretrained weights. We use the out-of-the-box Megadetector implementation since it is pretrained on millions of camera trap images and provides a strong initialisation for camera trap specific detection tasks. We then fine-tuned each model for 50 epochs using SGD with a batch size of 16. Training was carried out using an input image resolution of $416^2$ and an Intersection over Union (IoU) threshold of $0.5$ for non maximum suppression, at an initial learning rate of $1 \times 10^{-2}$ which was reduced by $10\%$ at $80\%$ and $90\%$ of the total training epochs. All ape detection models were evaluated using the commonly used object detection metrics: mean average precision (mAP), precision, recall and F1-scores. All metrics follow the open images standard~\citep{OpenImages2} and are considered in combination during evaluation. Performance is provided separately for small ($32^2$), medium ($96^2$) and large bounding boxes ($>96^2$), as per the COCO object detection standard, in addition to overall performance. 

\textbf{Performance}. Tab.~\ref{tab:ape_detection_results_test} shows that the fine-tuned Megadetector achieves the best mAP score overall and for large bounding boxes, although it is outperformed by the Swin Transformer and ResNet-101~(+Cascade R-CNN+SCM+TCM) on medium and small bounding boxes, respectively. This shows that in-domain pre-training of the feature extractor is valuable for fine-tuning since the Megadetector is the only model pretrained on a camera trap dataset, rather than the COCO dataset~\citep{lin2014microsoft}. Performance across the remaining metrics, precision, recall and F1-score, is dominated by the Swin Transformer, which shows the importance of modelling spatial dependencies for good detection performance.

\begin{figure*}[p!]
\centering
\includegraphics[width=0.99\linewidth]{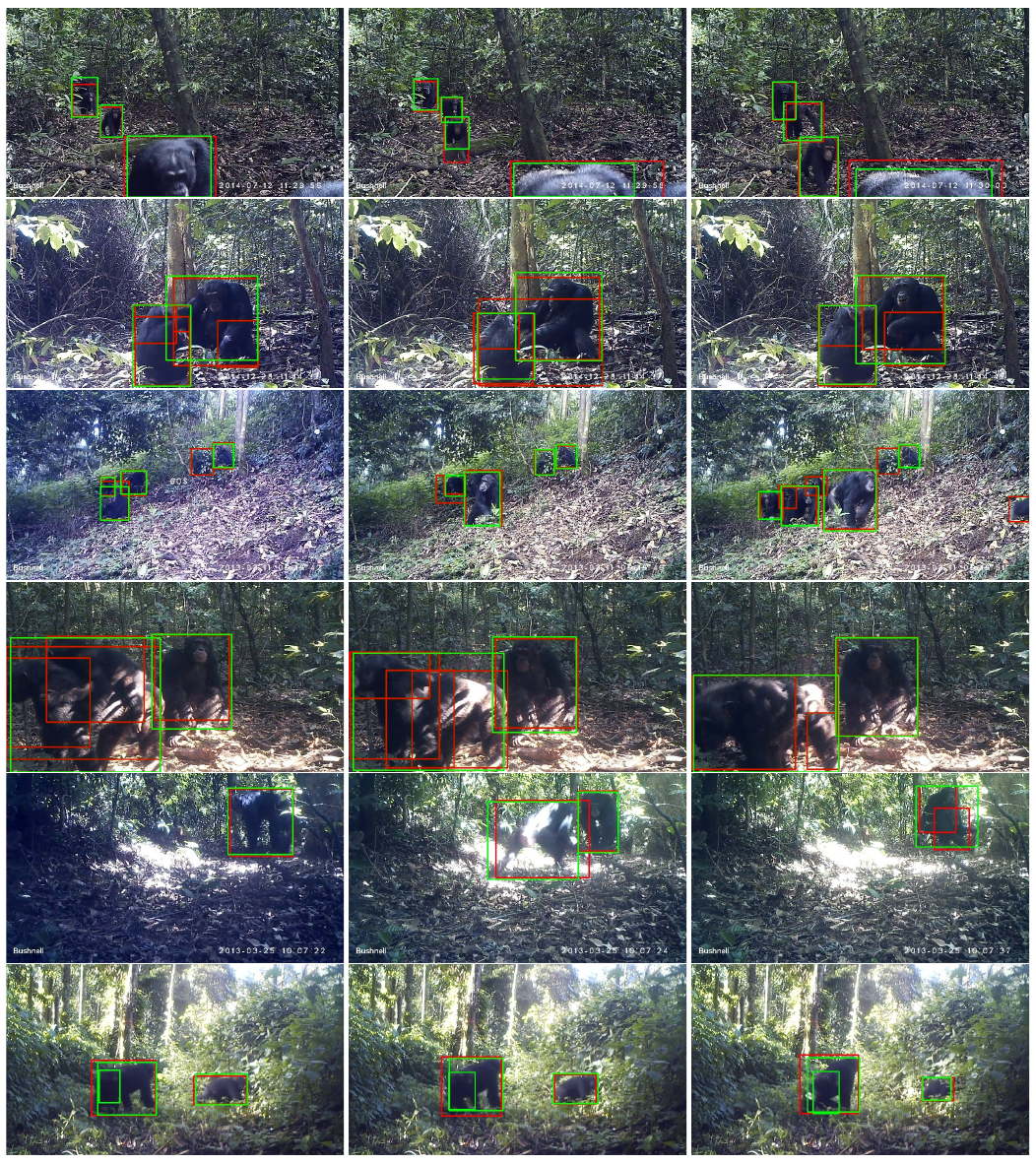}
\caption{\textbf{Megadetector detection examples}. A sequence of frames (along each row) extracted from 3 different videos. The ground truth bounding boxes (green) are shown alongside detections (red). The first sequence (\textbf{row 1}) shows successful detections. The second set of sequences (\textbf{row 2-4}) provide examples of false positive detections. The third set of sequences (\textbf{row 5-6}) provide examples of false negative detections.}
\label{fig:detection_examples}
\end{figure*}

The precision-recall (PR) curve displayed in Fig.~\ref{fig:detection_pr_curve} shows that most models maintain precision of more than 90\% ($P_{det} > 0.9$) at lower recall rates ($R_{det} < 0.80$), except ResNet-101~(+SCM+TCM) which falls below this at recall of 78\% ($R_{det}=0.78$). The fine-tuned Megadetector achieves consistently higher precision than other models for more than 84\% of cases ($R_{det}=0.84$), outperforming other models by 5\% ($P_{det}=0.05$) on average. However, at higher recall rates ($R_{det}>0.84$) ConvNeXt and SwinTransformer achieve higher precision, with the latter achieving marginally better performance. The ROC curve presented in Fig.~\ref{fig:detection_roc_curve} shows that VFNet and ResNet-101~(+SCM+TCM) achieve higher true positive rate than all other models at false positive rates less than 5\% ($FPR < 0.05$) and 40\% ($FPR < 0.40$), respectively. At higher false positive rates ConvNext and SwinTransformer are competitive with ResNet-101~(+SCM+TCM), with marginally better performance being established by ConvNeXt at very high false positive rates. Figure~\ref{fig:detection_examples} presents qualitative examples of success and failure cases for the best performing model.

\begin{figure}[!ht]
\centering
\includegraphics[width=\linewidth]{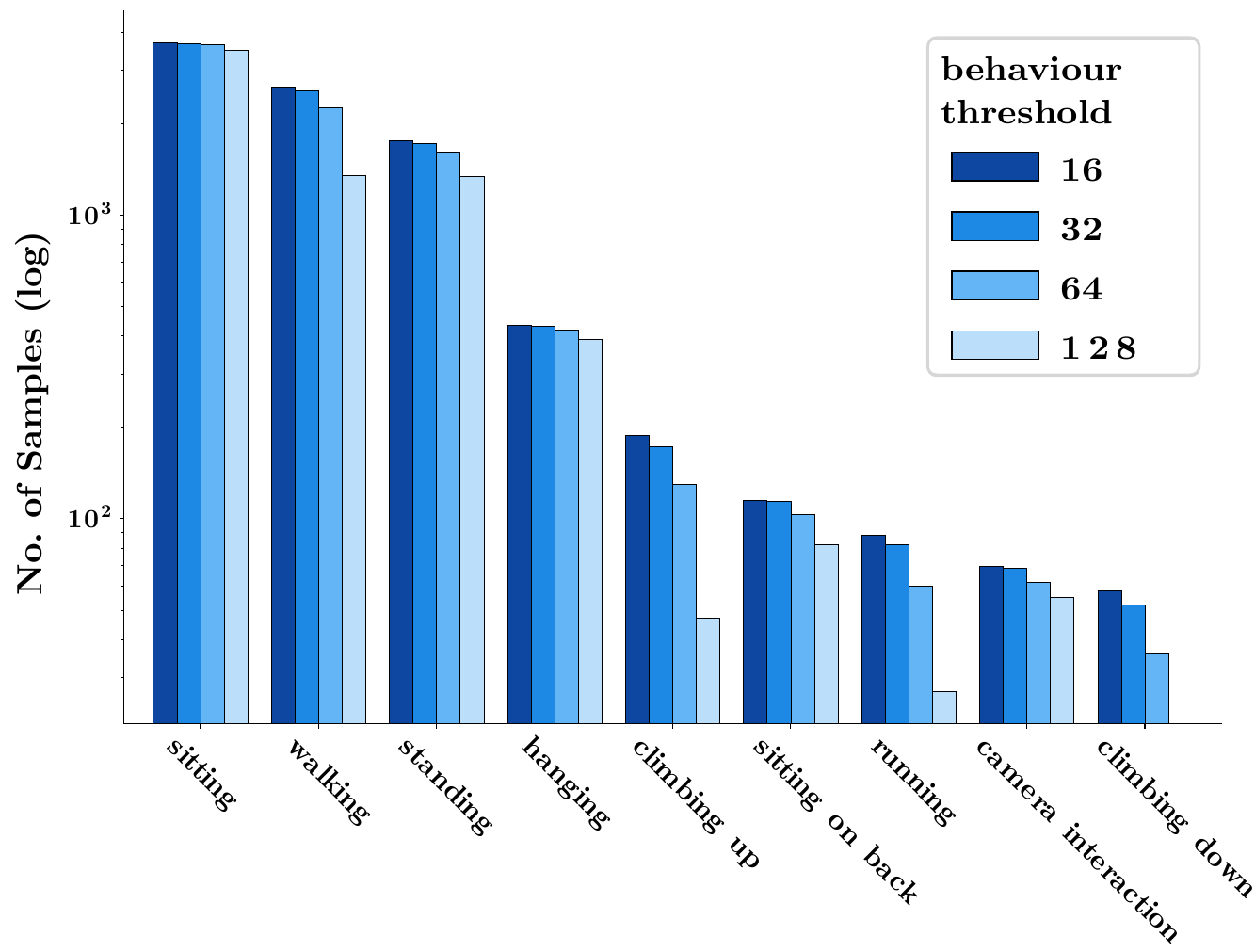}\vspace{-10pt}
\caption{\textbf{Per-class Distribution vs. Behavioural Thresholds}. \textmd{Distribution of each behavioural action class at various behavioural thresholds. Note that tail classes are effected more significantly by longer thresholds than head classes.}}
\label{fig:p500_thresh_dist}
\end{figure}

\begin{table*}[!hb]
\centering
\caption{\textbf{Behavioural Action Recognition Benchmarks.} \textmd{Behavioural action recognition performance on the PanAf500 dataset. Results are reported for X3D~\citep{feichtenhofer2020x3d}, I3D~\citep{carreira2017quo}, 3D ResNet-50~\citep{hara2017learning}, MViTV2~\citep{li2022mvitv2}, and TimeSformer~\citep{bertasius2021space} models. The highest scores for top-1 and average per-class accuracy are shown in bold.}}
\begin{tabular}{crrrrrrrr}
\toprule
\multirow{2}{*}{\textbf{Model}}
& \multicolumn{4}{c}{\textbf{Top-1 (\%)}}
& \multicolumn{4}{c}{\textbf{C-Avg (\%)}} \\
\cmidrule(r){2-5} \cmidrule(l){6-9}
 & \multicolumn{1}{c}{16} & \multicolumn{1}{c}{32} & \multicolumn{1}{c}{64} & \multicolumn{1}{c}{128} & \multicolumn{1}{c}{16} & \multicolumn{1}{c}{32} & \multicolumn{1}{c}{64} & \multicolumn{1}{c}{128}\\
\midrule
\multicolumn{1}{l}{X3D} & \textbf{80.00} & 80.04 & \textbf{79.40} & 74.24 & 50.35 & 56.10 & \textbf{53.02} & 40.89\\
\multicolumn{1}{l}{I3D} & 79.29 & 78.48 & 76.90 & 67.45 & 42.15 & 48.14 & 31.65 & 24.46\\
\multicolumn{1}{l}{3D ResNet-50} & 77.45 & 76.41 & 74.02 & 73.31 & \textbf{55.17} & 33.79 & 38.72 & 36.03\\
\multicolumn{1}{l}{MViTV2} & 78.31 & \textbf{81.09} & 79.29 & \textbf{79.19} & 40.45 & 54.91 & 48.28 & 41.11\\
\multicolumn{1}{l}{TimeSformer} & 78.53 & 79.45 & 79.26 & 78.18 & 45.05 & \textbf{56.38} & 48.27 & \textbf{41.10}\\
\bottomrule
\end{tabular}
\label{tab:behaviour_rec_results}
\end{table*}

\begin{figure*}[!ht]
\centering
\includegraphics[width=0.95\linewidth]{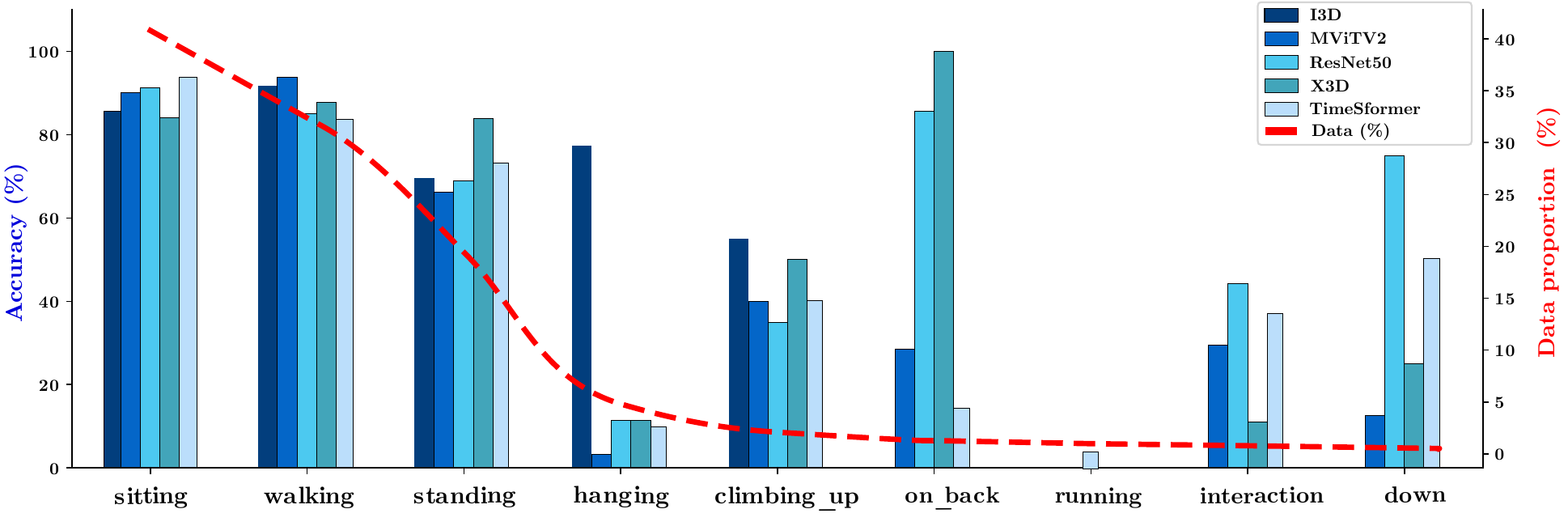}\vspace{-2pt}
\caption{\textmd{\textbf{Class-wise Performance vs. Proportion of Data.} The per-class accuracy for each behavioural action recognition model is plotted against the proportion of data for each class. All models consistently achieve strong performance on the head classes, whereas performance is variable across tail classes.}}
\label{fig:p500_per_class_acc_vs_samples}
\end{figure*}

\textbf{Behavioural Action Recognition}. We trained all models using the protocol established by~\citep{sakib2020visual}. During training we imposed a temporal behaviour threshold that ensures that only frame sequences in which a behaviour is exhibited for $t$ consecutive frames are utilised during training in order to retain well-defined behaviour instances. We then sub-sampled 16-frame sequences from clips that satisfy the behaviour threshold. The test threshold is always kept consistent ($t=16$). Fig.~\ref{fig:p500_thresh_dist} shows the effect of different behaviour thresholds on the number of clips available for each class. Higher behaviour thresholds have a more significant effect on minority/tail classes since they occur more sporadically. For example, there are no training clips available for the climbing down class where $t=128$. All models were initialised with feature extractors pre-trained on Kinetics-400~\citep{kay2017kinetics} and fine-tuned for 200 epochs using the Adam optimiser and a standard cross-entropy loss. We utilised a batch size of 32, momentum of 0.9 and performed linear warm-up followed by cosine annealing using an initial learning rate of $1\times10^{-5}$ that increases to $1\times10^{-4}$ over 20 epochs. All behavioural action recognition models were evaluated using average top-1 and average per-class accuracy (C-Avg).

\begin{figure*}[!hb]
\centering
\includegraphics[width=\linewidth]{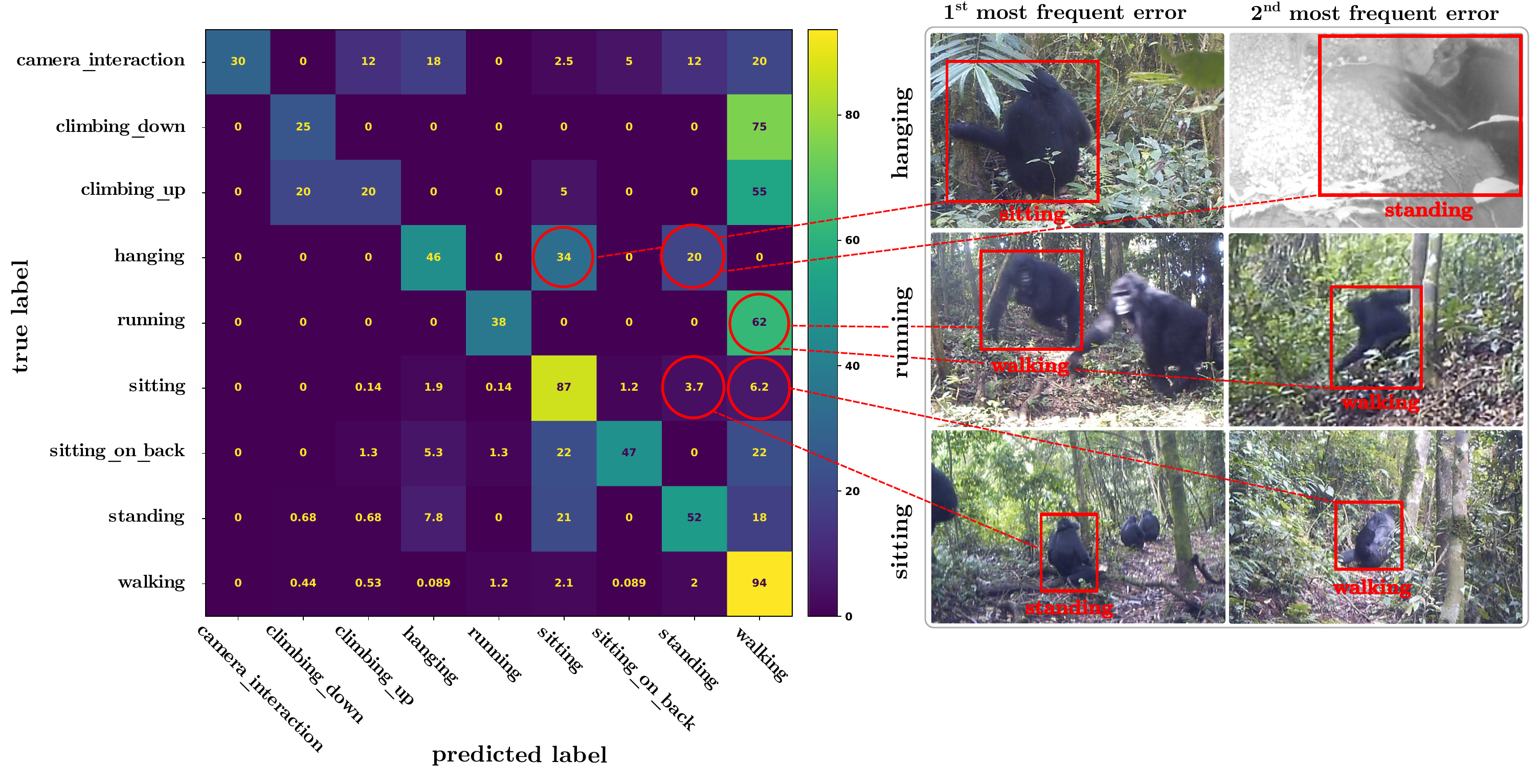}\vspace{-10pt}
\caption{\textmd{\textbf{Confusion Matrix \& Example Errors}. The confusion matrix (left) is shown alongside examples of mis-classified frames (right). For mis-classified examples, ground truth labels are shown on the y-axis (i.e., hanging, running, sitting) and examples of the classes most likely to be incorrectly predicted for the ground truth class are shown on the x-axis. Note that a high proportion of errors are due to predictions made in favour of majority classes.}}
\label{fig:p500_misclassifications}
\end{figure*}

\textbf{Performance}. Tab.~\ref{tab:behaviour_rec_results} shows the X3D model attains the best top-1 accuracy at behaviour thresholds $t=16$ and $t=64$, although similar performance is achieved by MViTV2 and TimeSformer for the latter threshold. It also achieves the best average per-class performance at $t=64$, while TimeSformer achieves the best performance at $t=32$ and $t=128$. 

The MVITV2 models realise the best top-1 accuracy at $t=32$ and $t=128$, although they do not achieve the best average per-class performance at any threshold. The 3D ResNet-50 achieves the best average per-class performance at $t=16$. When considering top-1 accuracy, model performance is competitive. At lower behavioural thresholds, i.e., $t=16$ and $t=32$, the difference in top-1 performance is 2.55\% and 4.68\%, respectively, between the best and worst performing models, although this increases to 5.38\% and 11.74\% at $t=64$ and $t=128$, respectively. There is greater variation in average per-class performance and it is rare that a model achieves the best performance across both metrics.

Although we observe strong performance with respect to top-1 accuracy, our models exhibit relatively poor average per-class performance. Fig.~\ref{fig:p500_per_class_acc_vs_samples} plots per-class performance against class frequency and shows that the average per-class performance is caused by poor performance on tail classes. The average per-class accuracy across all models for the head classes is 83.22\% while only 28.33\% is achieved for tail classes. There is significant variation in the performance of models; I3D performs well on hanging and climbing up but fails to classify any of the other classes correctly. Similarly, X3D performs extremely well on sitting on back but achieves poor results on the other classes. None of the models except for TimeSformer correctly classify any instances of running during testing. Fig.~\ref{fig:p500_misclassifications} presents the confusion matrix calculated on validation data alongside examples of misclassified instances.


\subsection{PanAf20K Dataset}

\begin{table*}[!hb]
\begin{center}
\caption{\textbf{Multi-label Behaviour Recognition Benchmarks}. Results are reported for I3D~\citep{carreira2017quo}, 3D ResNet-50~\citep{hara2017learning}, X3D~\citep{feichtenhofer2020x3d}, MViTV2~\citep{li2022mvitv2}, and TimeSformer~\citep{bertasius2021space} models with focal loss~\citep{cui2019class}, logit adjustment~\citep{menon2020long}, and focal loss with weight balancing~\citep{alshammari2022long}. The highest scores across all metrics are shown in bold.}
\begin{tabular}{crrrrrrr}
\toprule
\multirow{2}{*}{\textbf{Model}}
& \multicolumn{4}{c}{\textbf{mAP (\%)}}c
& \multicolumn{3}{c}{\textbf{Other (\%)}} \\
\cmidrule(r){2-5}\cmidrule(l){6-8}
 & \multicolumn{1}{c}{All} & \multicolumn{1}{c}{Head} & \multicolumn{1}{c}{Middle} & \multicolumn{1}{c}{Tail} & \multicolumn{1}{c}{Accuracy} & \multicolumn{1}{c}{Precision} & \multicolumn{1}{c}{Recall}\\
\midrule
\multicolumn{1}{l}{I3D} & 45.49 & 87.92 & 53.43 & 6.62 & 41.99 & 51.77 & 38.62 \\
\multicolumn{1}{l}{+FocalLoss} & 46.65 & 87.67 & 53.51 & 10.17 & 42.51 & 60.02 & 37.46 \\
\multicolumn{1}{l}{+LogitAdjustment} & 46.81 & 87.52 & 52.54 & 12.05 & 43.02 & 57.99 & 38.88 \\
\multicolumn{1}{l}{+WeightBalancing} & 46.41 & 88.41 & 51.91 & 11.07 & 41.93 & 57.36 & 35.62 \\
\hdashline
\multicolumn{1}{l}{3D ResNet-50} & 46.03 & 86.12 & 53.22 & 9.73 & 40.76 & 54.13 & 35.87 \\
\multicolumn{1}{l}{+FocalLoss} & 47.93 & 87.07 & \textbf{54.31} & 13.35 & \textbf{43.35} & 57.77 & 38.89 \\
\multicolumn{1}{l}{+LogitAdjustment} & 48.04 & 87.44 & 53.91 & 13.96 & 41.15 & 58.86 & 36.27 \\
\multicolumn{1}{l}{+WeightBalancing} & 46.68 & 87.09 & 54.06 & 9.90 & 41.54 & 54.05 & \textbf{40.62} \\
\hdashline
\multicolumn{1}{l}{X3D}~ & 46.06 & 87.32 & 52.95 & 9.36 & 42.70 & 49.26 & 39.25 \\
\multicolumn{1}{l}{+FocalLoss} & 47.19 & 89.26 & 52.75 & 11.75 & 41.67 & 50.93 & 36.99 \\
\multicolumn{1}{l}{+LogitAdjustment} & 47.85 & 88.58 & 53.06 & 13.77 & 42.64 & 59.18 & 36.94 \\
\multicolumn{1}{l}{+WeightBalancing} & 45.64 & 88.45 & 51.15 & 9.75 & 40.96 & 48.40 & 33.29 \\
\hdashline
\multicolumn{1}{l}{MViTV2} & 45.71 & 88.72 & 51.16 & 9.76 & 42.38 & 52.03 & 36.55 \\
\multicolumn{1}{l}{+FocalLoss} & 45.78 & \textbf{89.27} & 50.82 & 10.05 & 42.83 & 56.54 & 36.38 \\
\multicolumn{1}{l}{+LogitAdjustment} & 45.91 & 88.97 & 50.75 & 10.74 & 41.02 & 57.73 & 38.11 \\
\multicolumn{1}{l}{+WeightBalancing} & 45.36 & 88.58 & 49.82 & 10.59 & 42.44 & 49.77 & 36.20 \\
\hdashline
\multicolumn{1}{l}{TimeSformer} & 47.24 & 88.83 & 51.91 & 13.29 & 41.60 & 57.66 & 38.63 \\
\multicolumn{1}{l}{+FocalLoss} & 48.82 & 88.75 & 52.65 & 17.10 & 42.70 & 68.01 & 37.37 \\
\multicolumn{1}{l}{+LogitAdjustment} & \textbf{49.39} & 88.52 & 53.21 & \textbf{18.20} & 42.44 & \textbf{71.31} & 35.45 \\
\multicolumn{1}{l}{+WeightBalancing} & 48.17 & 87.98 & 51.81 & 16.78 & 41.86 & 61.16 & 39.36 \\
\bottomrule
\end{tabular}
\label{tab:multilabel_benchmarks}
\end{center}
\end{table*}

\textbf{Data Setup}. We generate train-val-test splits (70:10:20) using iterative stratification~\citep{sechidis2011stratification,2017arXiv170201460S}. During training, we uniformly sub-sample $t=16$ frames from each video, equating to $\sim$1 frame per second (i.e., a sample interval of 22.5 frames).

\textbf{Baseline Models.} To establish benchmark performance for multi-label behaviour recognition, we trained the X3D, I3D, 3D ResNet-50s, and MViTv2 models. All models were initialised with feature extractors pre-trained on Kinetics-400~\citep{kay2017kinetics} and fine-tuned for 200 epochs using the Adam optimiser. We utilised a batch size of 32, momentum of 0.9 and performed linear warm-up followed by cosine annealing using an initial learning rate of $1\times10^{-5}$ that increases to $1\times10^{-4}$ over 20 epochs. Models were evaluated using mAP, subset accuracy (i.e., exact match), precision and recall. Behaviour classes were grouped, based on class frequency, into head~($>10\%$), middle~($>1\%$) and tail~($<1\%$) segments, and mAP performance is reported for each segment. To address the long-tailed distribution, we substitute the standard loss for those calculated using long-tail recognition techniques. Specifically, we implement (i)~focal loss~\citep{cui2019class}~$L_{CB}$; (ii)~logit adjustment~\citep{menon2020long}~$L_{LA}$; and (iii)~focal loss with weight balancing via a MaxNorm constraint~\citep{alshammari2022long}.

\begin{figure*}[!ht]
\centering
\includegraphics[width=\linewidth]{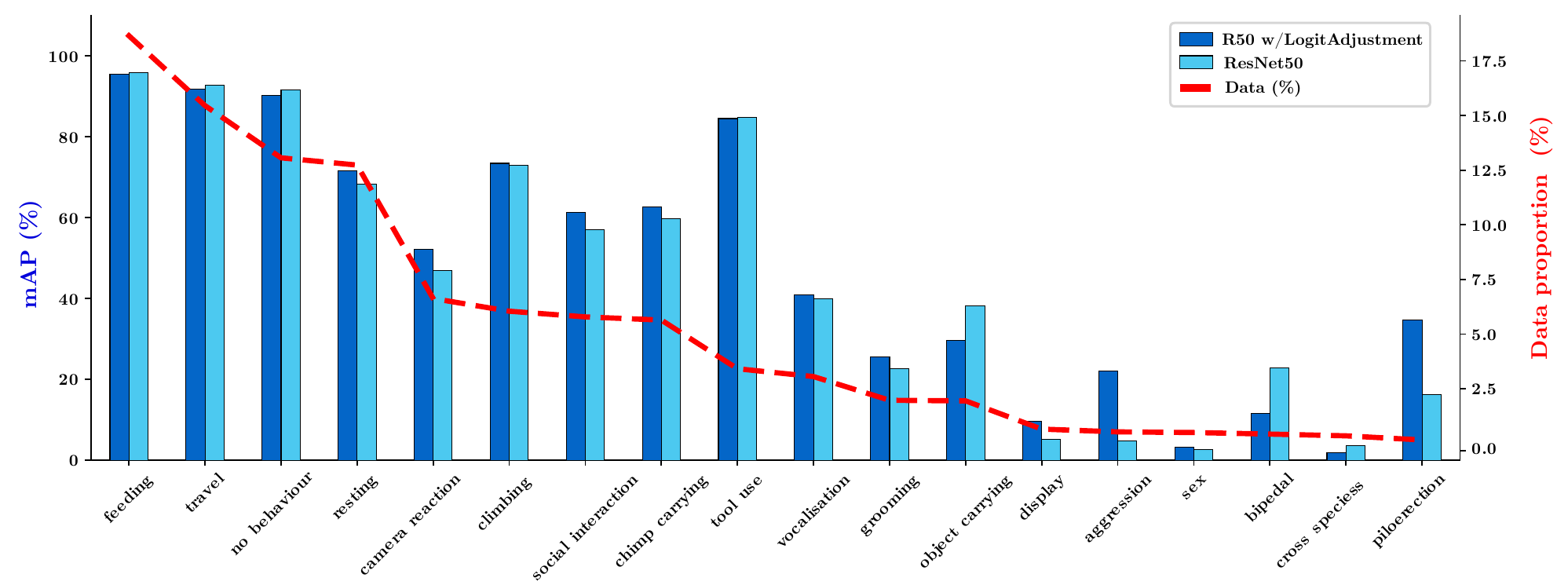}
\caption{\textmd{\textbf{Class-wise Accuracy vs. Proportion of Data.} The per-class average precision for the 3D ResNet-50~\citep{hara2017learning} and 3D ResNet-50~(+LogitAdjustment)~\citep{hara2017learning,menon2020long} models is plotted against the proportion of data for each class. In general, better model performance is achieved on classes with high data proportions and the ResNet-50 (+LogitAdjustment) model shows improved performance on middle and tail classes.}}
\label{fig:p20k_per_class_acc_vs_videos}
\end{figure*}

\textbf{Multi-label Behaviour Recognition.} As shown in Table~\ref{tab:multilabel_benchmarks}, performance is primarily dominated by the 3D ResNet-50 and TimeSformer models when coupled with the various long-tailed recognition techniques. The TimeSformer~(+LogitAdjustment) attains the highest mAP scores for both overall and tail classes, while the MViTV2~(+FocalLoss) and 3D ResNet-50~(+FocalLoss) demonstrate superior performance in terms of head and middle class mAP, respectively. The 3D ResNet-50~(+FocalLoss) and 3D ResNet-50~(+WeightBalancing) models achieve the best subset accuracy and recall, respectively, while the highest precision is realised by the TimeSformer~(+LogitAdjustment) model. Although the 3D ResNet-50 and TimeSformer models perform strongest, it should be noted that the difference in overall mAP across all models is small (i.e., 4.03\% between best and worst performing models).

\begin{figure*}[!ht]
\centering
\includegraphics[width=0.95\linewidth]{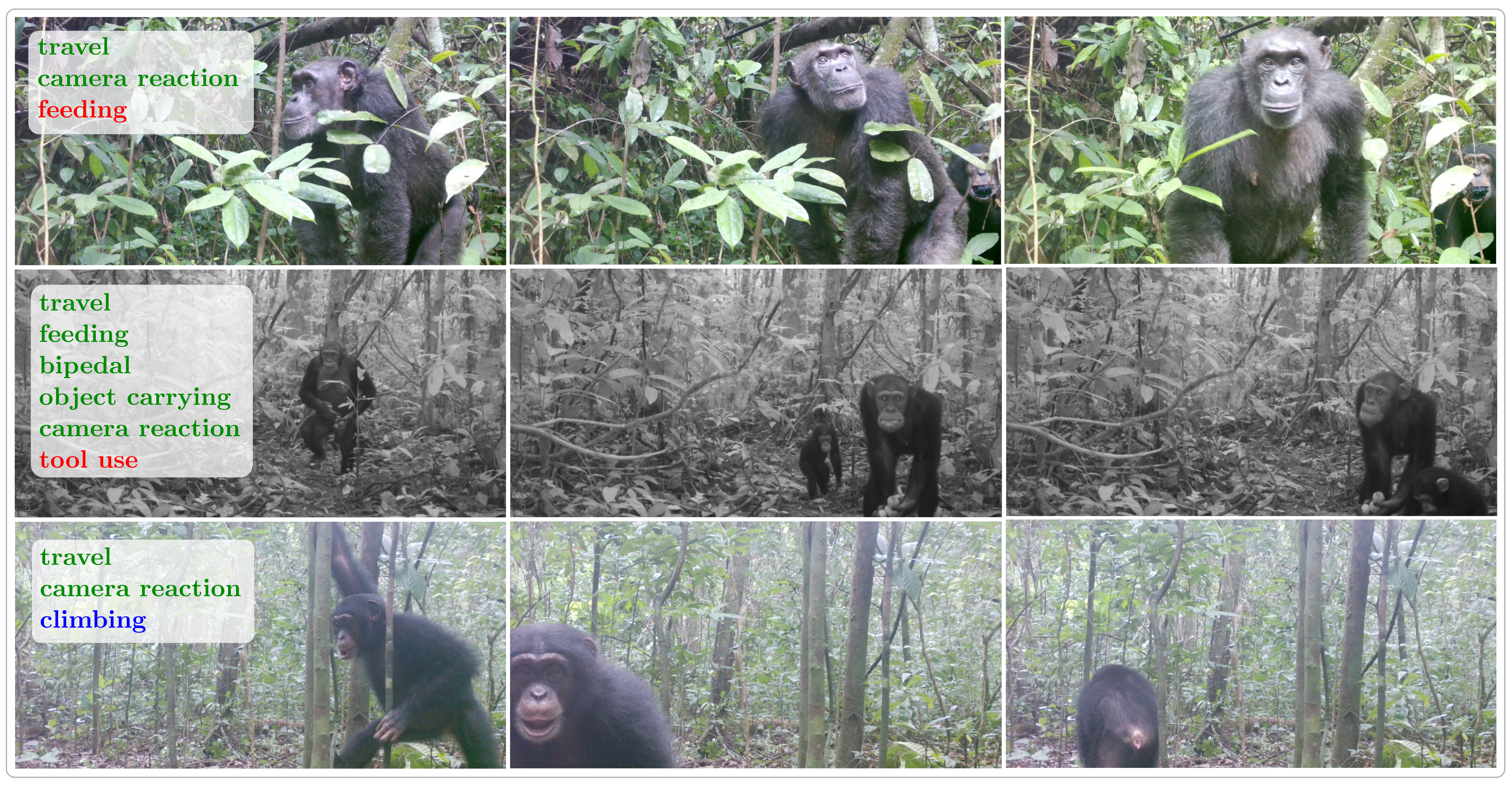}
\caption{\textmd{\textbf{Multi-label Errors.} Frames extracted from three videos exhibit success and failure cases of the 3D ResNet-50 model. Behaviour predictions are shown in light boxes of the first frame of each sequence; true positives are green, false positive are blue, and false negatives are red. In the first video (row 1), the model fails to classify feeding by the chimp visible in frames 1 and 2 whereas in the second video (row 2), it fails to classify tool use by the infant chimp in the final frame. Climbing is predicted incorrectly in the final video (row 3).}}
\label{fig:p20k_multi_label_errors}
\end{figure*}

As demonstrated by the head, middle and tail mAP scores, higher performance is achieved for more frequently occurring classes with performance deteriorating significantly for middle and tail classes. Across models, the average difference between head and middle, and middle and tail classes is 35.68~($\pm$1.88)\% and 40.55~($\pm$3.02)\%, respectively. The inclusion of long-tailed recognition techniques results in models that consistently attain higher tail class mAP performance than their standard counterparts (i.e., models that do not use long-tail recognition techniques). The logit adjustment technique consistently results in the best tail class mAP across models, whereas the focal loss results in the best performance on the middle classes for all models except the X3D model. None of the standard models achieve the best performance on any metric.

Fig.~\ref{fig:p20k_per_class_acc_vs_videos} plots per-class mAP performance of the 3D ResNet-50 and 3D ResNet-50(+LogitAdjustment) models against the per-class proportion of data. The best performance is observed for the three most commonly occurring classes (i.e., feeding, travel, and no behaviour) whereas the worst performance is obtained by the most infrequently occurring classes (i.e., display, aggression, sex, bipedal, and cross species interaction) with the exception of piloerection. It can also be observed that the ResNet-50(+LogitAdjustment) model outperforms its standard counterpart on the majority of middle and tail classes, although it is outperformed on tail classes. Examples of success and failure cases by the 3D ResNet-50 model are presented in Fig.~\ref{fig:p20k_multi_label_errors}.

\section{Discussion \& Future Work}\label{sec_discussion}

\textbf{Results}. The performance of current SOTA methods is not currently sufficient for facilitating the large-scale, automated behavioural monitoring required to support conservation efforts. The conclusions drawn in ecological studies rely on the highly accurate classification of all observed behaviours by expert primatologists. While the current methods achieve strong performance on head classes, relatively poor performance is observed for rare classes. Our results are consistent with recent work on similar datasets (i.e., AnimalKingdom~\citep{ng2022animal} and MammalNet~\citep{chen2023mammalnet}) which demonstrate the significance of the long-tailed distribution that naturally recorded
data exhibits~\citep{liu2019large}. Similar to~\citep{ng2022animal}, our experiments show that current long-tailed recognition techniques can help to improve performance on tail classes, although a large discrepancy between head and middle, and head and tail classes still exists. The extent of this performance gap (see Tab.~\ref{tab:multilabel_benchmarks}) emphasises the difficulty of tackling long-tailed distributions and highlights an important direction for future work~\citep{perrett2023}. Additionally, the near perfect performance at training time (i.e., $>95\%$ mAP) highlights the need for methods that can learn effectively from a minimal number of examples.

\textbf{Community Science \& Annotation}. Although behavioural annotations are provided by non-expert community scientists, several studies have shown the effectiveness of citizen scientists to perform complex data annotation tasks~\citep{danielsen2014multicountry,mccarthy2021chimpanzee} typically carried out by researchers (i.e., species classification, individual identification etc.). However, it should be noted that, as highlighted by~\citep{cox2012expert}, community scientists are more prone to errors relating to rare species. In the case of our dataset, this may translate to simple behaviours being identified correctly (e.g., feeding and tool use) whereas more nuanced or subtle behaviours (e.g., display and aggression) are missed or incorrectly interpreted, amongst other problems. This may occur despite the behaviour categories were predetermined by experts as suitable for non-expert annotation.

The dataset's rich annotations suit various computer vision tasks, despite key differences from other works. Unlike similar datasets~\citep{ng2022animal,chen2023mammalnet}, behaviours in the PanAf20K dataset are not temporally located within the video. However, the videos in our dataset are relatively short (i.e., 15 seconds) in contrast to the long form videos included in other datasets. Therefore, the time stamping of behaviour may be less significant considering it is possible to utilise entire videos, with a suitably fine-grained sample interval (i.e., 0.5-1 second), as input to standard action recognition models. With that being said, behaviours occur sporadically and chimpanzees are often only in frame for very short periods of time. Therefore, future work will consider augmenting the existing annotations with temporal localisation of actions. Moreover, while our dataset comprises a wide range of behaviour categories, many of them exhibit significant intra-class variation. In the context of ecological/primatological studies, this variation often necessitates the creation of separate ethograms for individual behaviours~\citep{nishida1999ethogram,zamma_matsusaka_2015}. For instance, within the tool use behaviour category, we find subcategories like nut cracking (utilizing rock, stone, or wood), termite fishing, and algae fishing. Similarly, within the camera reaction category, distinct subcategories include attraction, avoidance, and fixation. In future, we plan to extend the existing annotations to include more granular subcategories.

\textbf{Ethics Statement}. All data collection, including camera trapping, was done non-invasively, with no animal contact and no direct observation of the animals under study. Full research approval, data collection approval and research and sample permits of national ministries and protected area authorities were obtained in all countries of study. Sample and data export was also done with all necessary certificates, export and import permits. All work conformed to the relevant regulatory standards of the Max Planck Society, Germany. All community science work was undertaken according to the Zooniverse User Agreement and Privacy Policy. No experiments or data collection were undertaken with live animals.

\section{Conclusion}\label{sec:conclusion}
We present by-far the largest open-access video dataset of wild great apes with rich annotations and SOTA benchmarks. The dataset is directly suitable for visual AI training and model comparison. The size of the dataset and extent of labelling across $>$7M frames and $\sim$20K videos (lasting $>$80 hours) now offers the first comprehensive view of great ape populations and their behaviours to AI researchers. Task-specific annotations make the data suitable for a range of associated, challenging computer vision tasks (i.e, animal detection, tracking, and behaviour recognition) which can facilitate ecological analysis urgently required to support conservation efforts. We believe that given its immediate AI compatibility, scale, diversity, and accessibility, the PanAf20K dataset provides an unmatched opportunity for the many communities working in the ecological, biological, and computer vision domains to benchmark and expand great ape monitoring capabilities. We hope that this dataset can, ultimately, be a step towards better understanding and more effectively conserving these charismatic species.

\section*{Data availability}
All data and code will be made publicly available from the \href{https://obrookes.github.io/panaf.github.io/}{PanAf20K project website} upon publication and is available now upon request from the authors.

\backmatter

\bmhead{Acknowledgments}
We thank the Pan African Programme: ‘The Cultured Chimpanzee’ team and its collaborators for allowing the use of their data for this paper. We thank Amelie Pettrich, Antonio Buzharevski, Eva Martinez Garcia, Ivana Kirchmair, Sebastian Schütte, Linda Gerlach and Fabina Haas. We also thank management and support staff across all sites; specifically Yasmin Moebius, Geoffrey Muhanguzi, Martha Robbins, Henk Eshuis, Sergio Marrocoli and John Hart. Thanks to the team at https://www.chimpandsee.org particularly Briana Harder, Anja Landsmann, Laura K. Lynn, Zuzana Macháčková, Heidi Pfund, Kristeena Sigler and Jane Widness. The work that allowed for the collection of the dataset was funded by the Max Planck Society, Max Planck Society Innovation Fund, and Heinz L. Krekeler. In this respect we would like to thank: Ministre des Eaux et Forêts, Ministère de l'Enseignement supérieur et de la Recherche scientifique in Côte d’Ivoire; Institut Congolais pour la Conservation de la Nature, Ministère de la Recherche Scientifique in Democratic Republic of Congo; Forestry Development Authority in Liberia; Direction Des Eaux Et Forêts, Chasses Et Conservation Des Sols in Senegal; Makerere University Biological Field Station, Uganda National Council for Science and Technology, Uganda Wildlife Authority, National Forestry Authority in Uganda; National Institute for Forestry Development and Protected Area Management, Ministry of Agriculture and Forests, Ministry of Fisheries and Environment in Equatorial Guinea. This work was supported by the UKRI CDT in Interactive AI under grant EP/S022937/1.

\normalem
\bibliography{sn-bibliography.bib}

\end{document}